\documentclass[letterpaper, 10pt, conference]{ieeeconf}       % Use this line for a4 paper

\IEEEoverridecommandlockouts                              % This command is only needed if 
\usepackage{amsmath} % assumes amsmath package installed
\usepackage{mathtools} % loads amsmath
\usepackage{amsfonts}
\usepackage{xcolor}
\usepackage{graphicx}
\usepackage{booktabs}
\usepackage[caption=false,font=footnotesize]{subfig}
\usepackage{siunitx} % in your preamble
\usepackage[export]{adjustbox} % For easy alignment
\graphicspath{ {figures/} }
\usepackage{stfloats}
\usepackage{tikz}

\usepackage[hidelinks]{hyperref}
\usepackage[
style=ieee,
bibstyle=ieee,
isbn = false,
doi = true,
url = false,
eprint=true,
backend=biber]{biblatex}

\AtEveryBibitem{%
  \ifentrytype{dataset}
    {}
    {\clearfield{doi}}%
}
\AtEveryBibitem{\clearfield{address}}
\AtEveryBibitem{\clearfield{url}}
\AtEveryBibitem{\clearfield{month}}
\AtEveryBibitem{\clearlist{language}}
\AtEveryBibitem{\clearlist{location}}
\AtBeginBibliography{\footnotesize}

\DefineBibliographyStrings{english}{
  phdthesis   = Ph\adddot D\adddot\addspace thesis,
  mathesis    = Master's thesis,
}

\definecolor{blue}{HTML}{007BB9}
\definecolor{red}{HTML}{C81919}
\definecolor{green}{HTML}{006D2E}
\definecolor{purple}{HTML}{D700FF}
\definecolor{yellow}{HTML}{DFFF00}

\newtheorem{definition}{Definition}

\title{\LARGE \bf
Visualizing Impedance Control in Augmented Reality for Teleoperation: Design and User Evaluation
}

\author{Gijs van den Brandt, Femke van Beek, Elena Torta% <-this % stops a space
\thanks{The authors are with the Department of Mechanical Engineering, Eindhoven University of Technology, The Netherlands. Corresponding email:
        {\tt\small a.a.h.m.v.d.brandt@tue.nl}}%
}

\begin{document}

\maketitle
\thispagestyle{empty}
\pagestyle{empty}
%%%%%%%%%%%%%%%%%%%%%%%%%%%%%%%%%%%%%%%%%%%%%%%%%%%%%%%%%%%%%%%%%%%%%%%%%%%%%%%%
\begin{abstract}
Teleoperation for contact-rich manipulation remains challenging, especially when using low-cost, motion-only interfaces that provide no haptic feedback. Virtual reality controllers enable intuitive motion control but do not allow operators to directly perceive or regulate contact forces, limiting task performance. To address this, we propose an augmented reality (AR) visualization of the impedance controller's target pose and its displacement from each robot end effector. This visualization conveys the forces generated by the controller, providing operators with intuitive, real-time feedback without expensive haptic hardware. We evaluate the design in a dual-arm manipulation study with 17 participants who repeatedly reposition a box with and without the AR visualization. Results show that AR visualization reduces completion time by ~24\% for force-critical lifting tasks, with no significant effect on sliding tasks where precise force control is less critical. These findings indicate that making the impedance target visible through AR is a viable approach to improve human-robot interaction for contact-rich teleoperation.
\end{abstract}

\maketitle
\thispagestyle{empty}
\pagestyle{empty}
% \color{red}

%%%%%%% Uncomment this for arxiv version %%%%%%%%%%%%%%%%%%%%%%%%%%%%%%%%%%%%%%
\begin{tikzpicture}[remember picture,overlay]
\node[anchor=south,yshift=10pt] at (current page.south) {%
\fbox{%
\parbox{\dimexpr0.55\textwidth-\fboxsep-\fboxrule\relax}{%
\centering
\footnotesize
This work has been accepted for publication at the 35th IEEE International Conference on Robot and Human Interactive Communication (RO-MAN 2026).
}%
}%
};
\end{tikzpicture}

% \color{black}
\section{INTRODUCTION}
\begin{figure}[]
    \vspace{4pt}
    \centering
    % \rule{0.8\linewidth}{0.3\textwidth} % width x height black rectangle
    \includegraphics[width=\linewidth]{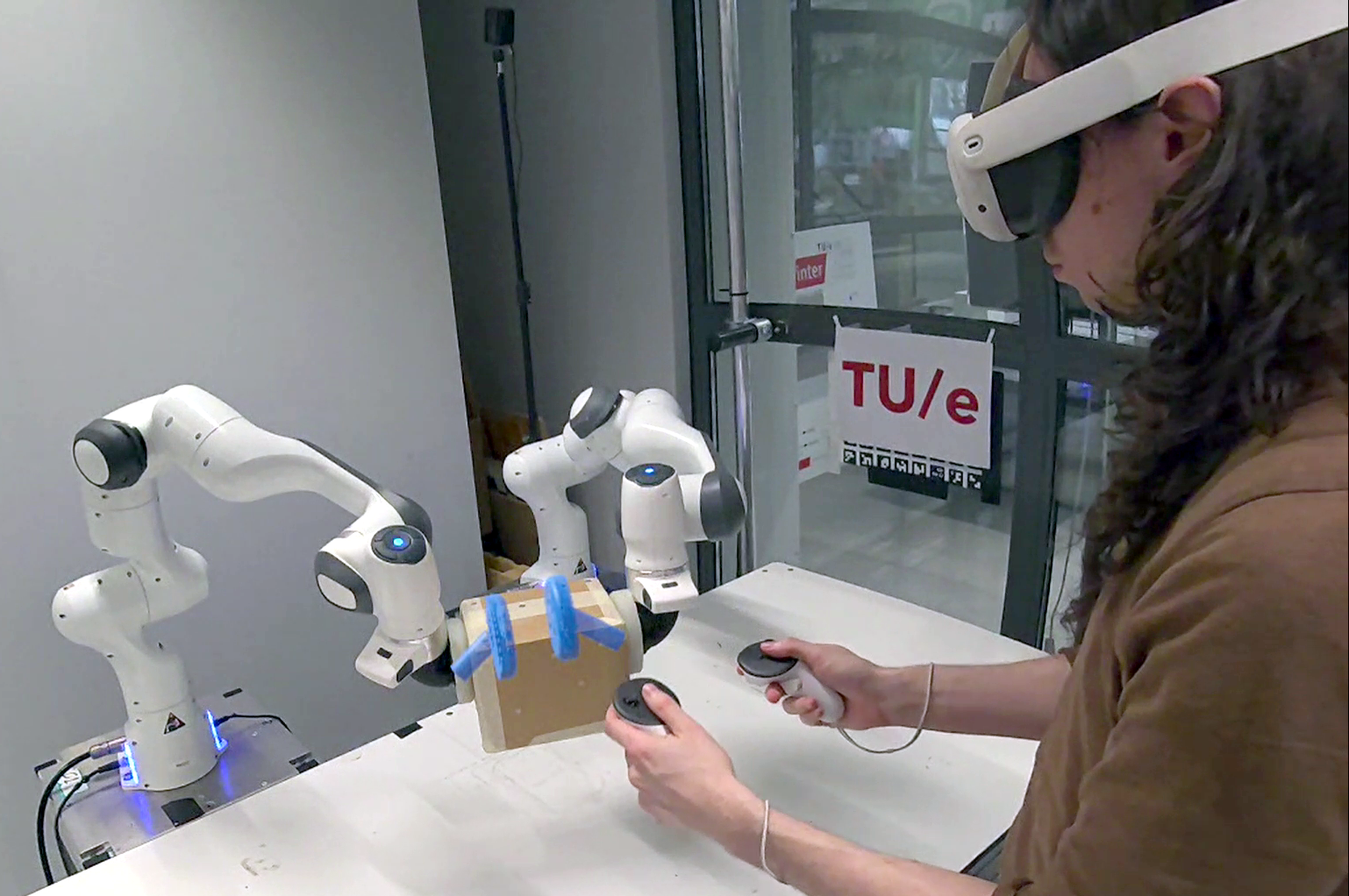}
    \caption{Example of contact-rich teleoperation. A human operator collaborates with two robot arms to lift a box; a task requiring careful regulation of contact forces. The operator uses VR motion-tracking controllers with an impedance controller to command the end effectors. Our proposed AR visualization displays the impedance controller's target pose (blue disks) and the position tracking offset (blue lines), providing intuitive visual feedback that correlates with the generated forces. A video of participants operating the system is available at \url{https://youtu.be/Dj36tJmo-6Q}.}
    \label{fig:scenario_a}
\end{figure}

Robotic manipulation in unstructured, contact-rich environments remains a major challenge for autonomous systems~\cite{Billard2019}. Tasks such as repositioning objects with dual-arm robots, as illustrated in Fig.~\ref{fig:scenario_a}, require sustained multi-contact interactions and careful coordination of motion and forces under uncertainty. Because reliably handling such interactions autonomously is difficult, human operators remain essential for performing these tasks via teleoperation, both in remote applications~\cite{Darvish2023} and for collecting demonstrations for learning-based methods~\cite{Si2021}.

Making teleoperation widely accessible requires low-cost, intuitive input interfaces.  As an alternative to leader-follower setups \cite{Zhao2023a} that are robot-specific and expensive, consumer virtual reality (VR) controllers with 6-DoF motion tracking have become a popular choice due to their affordability and flexibility~\cite{Wonsick2020}. They enable relatively intuitive control of robot end effectors and have been successfully applied in scenarios where grasping can be simplified to binary gripper commands~\cite{Whitney2018,Naceri2019}. However, many manipulation tasks are contact-rich and require precise regulation of interaction forces. VR controllers sense motion but provide no force input or feedback, limiting operator ability to manage forces during contact.

To address the lack of force input in teleoperation with VR controllers, impedance control can be used \cite{VanSteen2024, Zhou2026a}. Impedance control generates interaction forces by emulating a virtual spring and damper between the robot and the commanded pose~\cite{Hogan1985}. An operator can command a pose beyond a surface so that the virtual spring deflection produces contact forces. However, impedance control alone does not provide haptic feedback. Prior research indicates that force feedback improves teleoperation~\cite{Talasaz2017}, but VR controllers do not allow operators to directly perceive the forces generated by an impedance controller, nor the underlying virtual target pose. This lack of transparency complicates teleoperation.

In this work, we address the gap in transparency during teleoperation by introducing an augmented reality (AR) visualization of the otherwise hidden target pose of the impedance controller. Our interface overlays the target pose and the target position offset directly onto the operator's view, providing intuitive, real-time feedback that correlates with the forces generated by the controller.

We evaluated the proposed visualization in a user study involving dual-arm manipulation tasks. Participants completed object repositioning tasks under conditions with and without the augmented reality visualization. A pilot study ($n=10$) was performed to refine the design and experimental protocol, followed by the main study ($n=17$). The results indicate that visualizing the impedance target significantly speeds up lifting tasks that require careful force regulation, while having negligible effect on sliding tasks.

\section{RELATED WORK}
Dual-arm manipulation with non-articulated end effectors has been explored in several works~\cite{Simeonov2021a,Ozcan2025}. This scenario exemplifies tasks that involve regrasping objects and maintaining sustained contact through careful regulation of contact forces. Humans are relatively good at such manipulation, and a suitable teleoperation interface can help transfer this skill to robots. Haptic interfaces are most effective for these contact-rich tasks but are more expensive than VR devices. Recent efforts, such as ALOHA~\cite{Zhao2023a}, have somewhat reduced the cost of haptic interfaces, but these systems are robot-specific and limit operator mobility.

Commercial VR interfaces remain cheaper and offer greater flexibility compared to haptic devices, allowing operators to move freely and control a variety of robots; however, they lack direct force feedback. Most prior work on VR-based teleoperation has focused on robots with grippers performing tasks that do not require precise force control (e.g.~\cite{Whitney2018,Naceri2019}).
Van Steen et al.~\cite{VanSteen2024} used a VR interface with impedance control to teleoperate two robots with flat-faced end effectors, but their application was limited to a simple lifting task without regrasping, performed by a trained operator. Our work aims to enhance VR-based teleoperation in similar scenarios by providing visual feedback on the impedance controller through augmented reality.

Augmented reality has been widely explored to enhance VR-based teleoperation by improving operator awareness and control. Many works focus on spatial scene understanding, such as improving depth perception through depth-based color coding~\cite{Kim2025a} or augmenting the user's view with additional camera perspectives for fine-grained manipulation~\cite{Iyer2025}. Visual force feedback specifically has also been investigated, though in applications different from ours. Talasaz et al.~\cite{Talasaz2017} visualized force magnitude, but not direction, using a bar graph in robotic surgery. Chan et al.~\cite{Chan2022} decomposed force into three orthogonal directions and visualized them with corresponding arrows for a task that resembles polishing. Zhang et al.~\cite{Zhang2018} used a three-dimensional arrow to represent human-intended control for a dual-arm robot with grippers. While this idea of visualizing intended control aligns with our own, we study a gripperless application, we extend visualization beyond a single arrow by representing the full 6-DoF target, and we explicitly couple this visualization to an impedance controller.

\section{IMPEDANCE CONTROL AND VISUALIZATION}
This section describes the impedance control framework underlying our teleoperation interface and the corresponding augmented reality visualization. First, we explain impedance control conceptually, highlighting how users can influence interaction forces by moving a virtual target. Next, we describe and motivate the design of the visualization. Finally, we describe its integration into a VR-based teleoperation system.
\begin{figure}[b]
        \centering
        \includegraphics[]{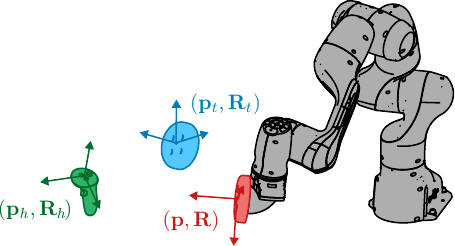}
        \caption{Schematic drawing of a robot arm (grey), the current end-effector pose (red), the target pose (blue) and the handheld VR controller (green) with their corresponding coordinate frames. The target pose is a digital entity and cannot be seen by the naked eye.}
        \label{fig:frames}
\end{figure}
\subsection{Concept of Impedance Control}

In contact-rich telemanipulation, the operator must control not only the end-effector position but also the force applied to the environment. The relationship between motion and force is characterized by mechanical impedance. For intuitive teleoperation, this impedance should feel natural and transparent to the user. A common approach is to connect the current and target end-effector poses through an emulated spring-damper~\cite{VanSteen2024,Zhou2026a,Hogan1985}. Conceptually, the robot behaves as if connected to the target by a rubber band: moving the target stretches the band, generating a force that drives the end effector to follow. This force is directed at the target and is larger if the target is further away. If the target is obstructed by an object, the end effector exerts force on the object instead. This rubber band-like behavior is  intuitive to humans and, therefore, a suitable interface for teleoperation. Definition~\ref{def:impedance} provides the formalization of the impedance controller that matches this conceptual description, describing how the joint torques required for controlling the robot are derived from the current and target end-effector poses. Fig. \ref{fig:frames} illustrates these poses.

\begin{definition}[Impedance controller]\label{def:impedance}
Consider an end effector with current position $\mathbf{p}\in\mathbb{R}^3$, orientation $\mathbf{R}\in SO(3)$, translational velocity $\mathbf{v}\in\mathbb{R}^3$, and angular velocity $\boldsymbol{\omega}\in\mathbb{R}^3$.
Let $\mathbf{p}_t$, $\mathbf{R}_t$, $\mathbf{v}_t$, and $\boldsymbol{\omega}_t$ denote the target position, rotation, and velocities.
The impedance wrench $\mathbf{F} \in \mathbb{R}^6$ is defined as
\begin{equation}
\mathbf{F}
=
\mathbf{K}
\begin{bmatrix}
\mathbf{p}_t - \mathbf{p} \\
\boldsymbol{\phi}(\mathbf{R}^\top \mathbf{R}_t)
\end{bmatrix}
+
\mathbf{D}
\begin{bmatrix}
\mathbf{v}_t - \mathbf{v} \\
\boldsymbol{\omega}_t - \boldsymbol{\omega}
\end{bmatrix},
\label{eq:cartesian_impedance}
\end{equation}
where $\mathbf{K}\in\mathbb{R}^{6\times6}$ is the diagonal stiffness matrix,
$\mathbf{D}\in\mathbb{R}^{6\times6}$ is the damping matrix, and
$\boldsymbol{\phi}(\mathbf{R}^\top \mathbf{R}_t)\in\mathbb{R}^3$ denotes the rotation vector
associated with the relative rotation from $\mathbf{R}$ to $\mathbf{R}_t$.
The corresponding joint-space torque command $\boldsymbol{\tau}\in\mathbb{R}^n$ for all $n$ joints
is obtained via the Jacobian transpose,
\begin{equation}
\boldsymbol{\tau} = \mathbf{J}^\top\mathbf{F},
\end{equation}
where $\mathbf{J}\in\mathbb{R}^{6\times n}$ is the end-effector Jacobian.
\end{definition}

\begin{figure*}[t]
    \vspace{4pt}
    \vspace{-10pt} % ensures no extra gap
    \centering
    \subfloat[]{\includegraphics[width=0.25\textwidth]{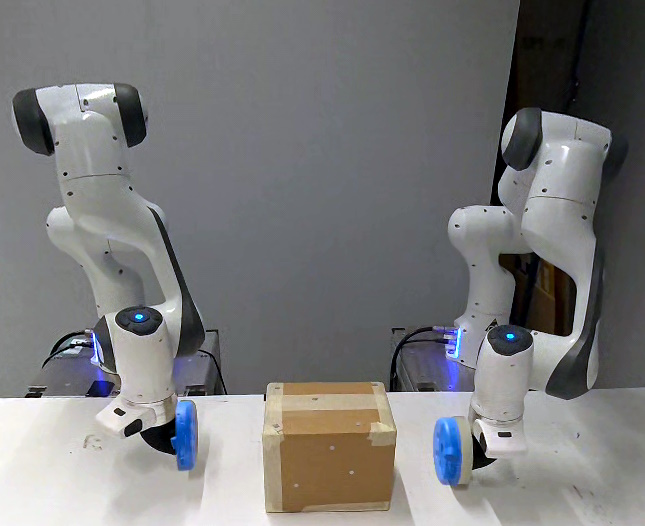}}%
    \hfill
    \subfloat[]{\includegraphics[width=0.25\textwidth]{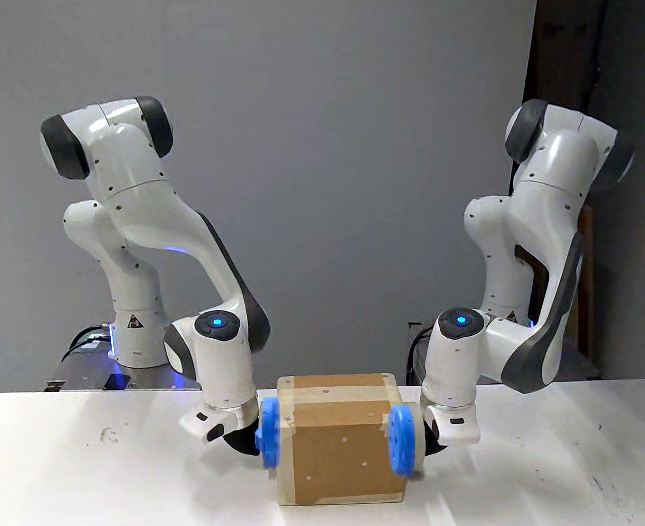}}%
    \hfill
    \subfloat[]{\includegraphics[width=0.25\textwidth]{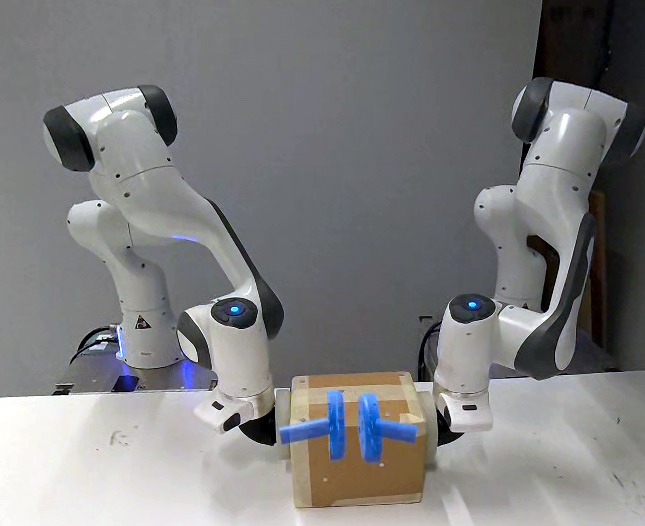}}%
    \hfill
    \subfloat[]{\includegraphics[width=0.25\textwidth]{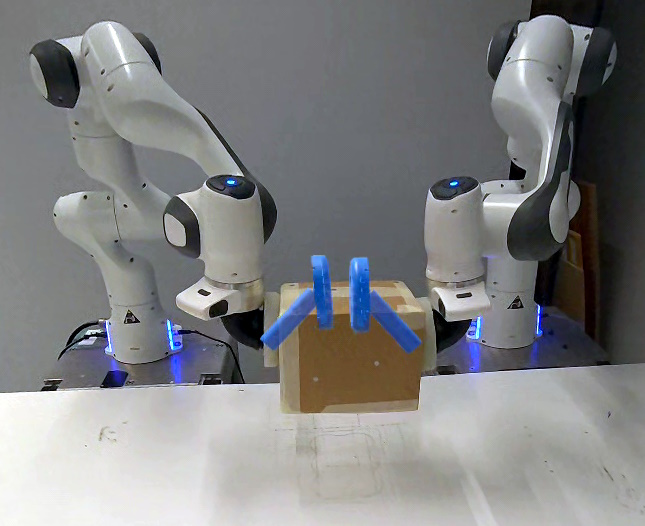}}

    \caption{Design of our impedance visualization shown from the perspective of an operator lifting an object. The blue disks represent the 6-DoF target poses of the end effectors, while the blue lines indicate target offset in translation, which correlate with the virtual forces applied by the controller. Four consecutive snapshots show the task progression: (a) no contact, (b) initial contact, (c) clamping the box, and (d) lifting it. Without this visualization, users cannot directly perceive the difference between (b) and (c), making force regulation difficult.}
    \label{fig:design}
\end{figure*}

\subsection{AR Visualization of the Impedance Target}
The impedance controller from Definition~\ref{def:impedance} enables operators to regulate both motion and contact forces through specification of an end-effector target. In practice, however, controlling this virtual target is difficult, as it is not directly observable. The robot itself provides an unreliable indication: it lags behind the target, which can cause overshooting, and it may fail to converge when the target lies outside the reachable workspace or is blocked by an object. Without direct feedback on the target pose, and by extension the contact forces, precise teleoperation becomes challenging.

To address this problem, we propose an augmented reality (AR) visualization that provides operators with direct, real-time feedback on the impedance target. The design, illustrated in Fig.~\ref{fig:design}, consists of two main elements:

\subsubsection{Target disk} A blue copy of the end-effector geometry represents the 6-DoF target pose. While poses are often represented with coordinate frames composed of three orthogonal axes, we believe that a copy of the end effector provides a more intuitive mapping of rotation to the physical end effector. This visualization is similar to what Smith and Kennedy~\cite{Smith2026} proposed for a velocity-controlled gripper.

\subsubsection{Target Offset Line} A blue line connects the end effector to the target. This line represents the difference between the current and target end-effector position, which contributes to the virtual force applied by the impedance controller through the stiffness following Definition \ref{def:impedance}. The line also correlates each target disk with the correct robot.

The disk and line are translucent and overlaid on the real-world video pass-through, enabling simultaneous perception of the environment and controller state. Although this paper evaluates the design with disk-shaped end effectors, the target offset line generalizes directly to other robots, while the target shape can be adapted to any end-effector geometry.

\subsection{System Implementation}
The system consisted of two Franka Emika Panda robots, spaced 90~cm apart and equipped with custom silicone end effectors introduced in~\cite{VanSteen2024}. The end effectors were mounted at a 90° angle, since this appeared to improve coverage of the task-relevant workspace during our pilot study.

Users wore a Meta Quest 3 headset running a custom Unity app, which provided a video pass-through of the real environment augmented with impedance visualizations. Handheld Meta Quest 3 controllers specified the impedance target. While a clutch button was pressed, the target followed the controller motion. Releasing the clutch decoupled the hand and target, allowing hand repositioning without affecting the robot. This mapping between the hand and the target is formalized in Definition \ref{def:clutch_teleop}. Similar approaches appear in \cite{Naceri2019} and \cite{Li2026a}. The resulting target pose and velocity were sent from the Unity app to the robot PC.

On the robot PC, a ROS node ran a modified version of the Cartesian impedance controller from the franka\_ros package\footnote{\url{https://github.com/frankarobotics/franka_ros}}. This modification avoids unnecessary damping by setting the target end-effector velocity to match the operator's hand rather than setting it to zero. Tracking only the pose but not the velocity results in slow convergence and thus unresponsive teleoperation, especially since we chose a relatively low translational stiffness of 200~N/m to prevent excessive contact forces. As a comparison, for a setup with grippers, Zhou et al.~\cite{Zhou2026a} used the unmodified Cartesian impedance controller meaning they did not track hand velocity; however, they used a higher stiffness of 600 N/m.

\begin{definition}[Hand-target mapping]\label{def:clutch_teleop}
Consider a handheld VR controller with pose $(\mathbf{p}_h, \mathbf{R}_h)$ and velocity $(\mathbf{v}_h, \boldsymbol{\omega}_h)$.  
Let $b \in \{0,1\}$ denote the clutch state (button pressed $b=1$, released $b=0$).  
Define the anchor poses $(\mathbf{p}_h^\mathrm{a}, \mathbf{R}_h^\mathrm{a})$ and $(\mathbf{p}_t^\mathrm{a}, \mathbf{R}_t^\mathrm{a})$ as the controller and target poses recorded at the most recent change of clutch state. The target pose $(\mathbf{p}_t, \mathbf{R}_t)$ and target velocity $(\mathbf{v}_t, \boldsymbol{\omega}_t)$ are then
\begin{align}
\mathbf{p}_t = \mathbf{p}_t^\mathrm{a} + b \, (\mathbf{p}_h - \mathbf{p}_h^\mathrm{a})&,
\qquad
\mathbf{R}_t =  \mathbf{R}_t^\mathrm{a} \,( \mathbf{R}_h^{\mathrm{a}\top} \mathbf{R}_h)^b ,\\
% \end{equation}
% \begin{equation}
\mathbf{v}_t = b \, \mathbf{v}_h&,
\qquad
\boldsymbol{\omega}_t = b \, \boldsymbol{\omega}_h.
\end{align}
Hence, if $b=1$, the target pose follows the controller pose relative to the respective anchors, and the target velocity matches the controller velocity.  
If $b=0$, the target pose remains fixed at the anchor from the most recent clutch release, and the target velocity is zero.
\end{definition}

\section{USER STUDY}
We conducted a user study to evaluate the effect of our augmented reality visualization of the impedance controller on dual-arm telemanipulation tasks. The study employed a within-subjects, repeated-measures 2×2 design, with two independent variables: interface condition (visualization and no visualization) and task type (lifting and sliding). Participants completed eight targets per task type under each interface condition, for a total of 32 recorded targets per participant. This study was approved by Eindhoven University of Technology's Ethical Review Board (\#ERB-445). The following subsections describe the manipulation tasks, experimental protocol, measures, and participant demographics.

\subsection{Manipulation Tasks}
\begin{figure}[t]
    \vspace{4pt}
    \vspace{-10pt} % ensures no extra gap
    \subfloat{\includegraphics[width=0.50\linewidth]{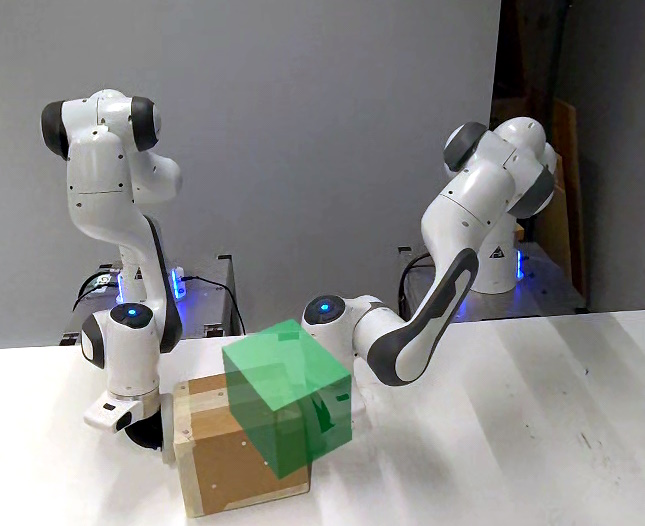}}%
    \hfill
    \subfloat{\includegraphics[width=0.50\linewidth]{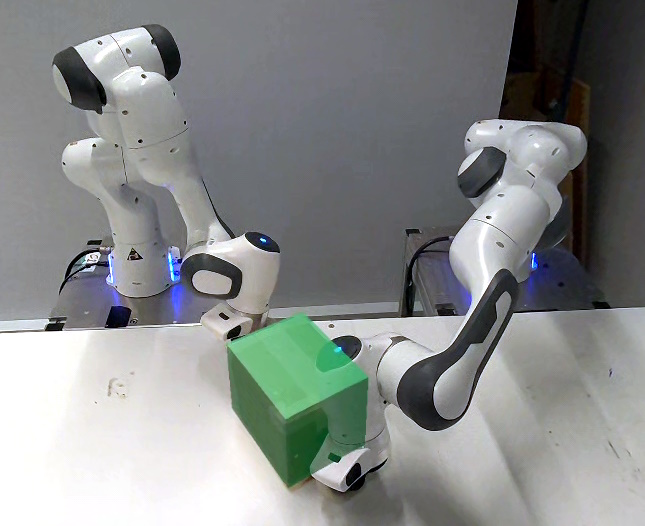}}%
    \vspace{-10pt} % ensures no extra gap
    \subfloat{\includegraphics[width=0.50\linewidth]{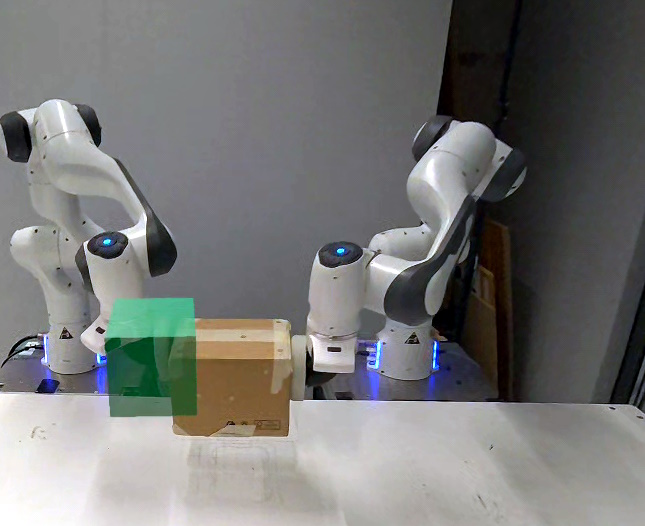}}%
    \hfill
    \subfloat{\includegraphics[width=0.50\linewidth]{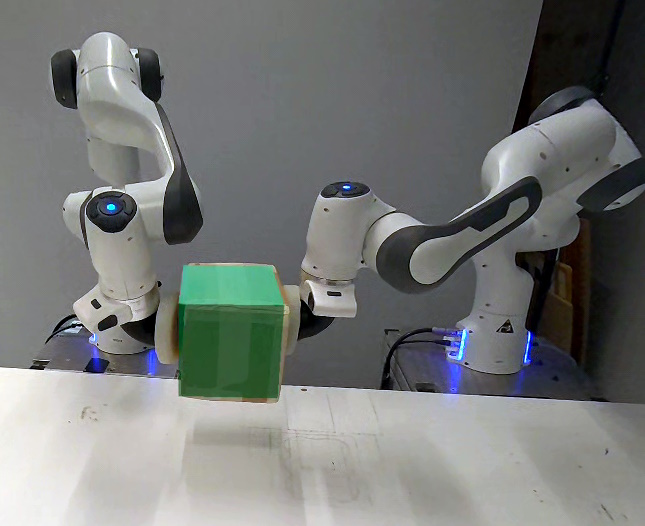}}
    \caption{Examples of the manipulation tasks: sliding (top) and lifting (bottom). The green cuboid, displayed in augmented reality, indicates the target location for the box. Images on the left show when the target first appeared, while images on the right show the moment of target completion.}
    \label{fig:task}
\end{figure}

Participants manipulated a rectangular box using two robotic arms, moving it from an initial pose to a target pose. Two types of tasks were performed: sliding and lifting (Fig. \ref{fig:task}). Sliding tasks allowed the box to remain in contact with the table, while lifting tasks involved raising the box off the table, requiring careful force control to avoid dropping. Targets varied in 3D position and in rotation around the vertical axis. The variance in rotation required participants to occasionally place the box down to get a different grasp. 

The box measured 16 $\times$ 16 $\times$ 20~cm and had a mass of 635~g. An OptiTrack motion capture system was used to verify whether the box pose met the target completion criteria formulated in Definition~\ref{def:task_completion}. Targets were presented sequentially, with a new target appearing immediately after the previous target was reached. Targets were displayed as a green cuboid through the VR headset, as illustrated in Fig.~\ref{fig:task}. This target visualization in AR combined with box tracking allowed for larger flexibility in targets than using a physical fixture (e.g., \cite{Falcone2025}).

\begin{definition}[Target Completion Criteria]\label{def:task_completion}
A target was considered complete when the box pose was sufficiently close in both position and orientation. Specifically, the Euclidean distance between the box position $\mathbf{p}_{\text{box}}$ and the target position $\mathbf{p}_{\text{target}}$ had to be smaller than $\delta_p = 30~\text{mm}$:
\begin{equation}
\|\mathbf{p}_{\text{box}} - \mathbf{p}_{\text{target}}\|_2 < \delta_p.
\end{equation}
The rotational alignment between the box orientation $\mathbf{R}_{\text{box}}$ and the target orientation $\mathbf{R}_{\text{target}}$ had to be within $\delta_\theta = 0.4~\text{rad}$. To account for the box's $180^\circ$ symmetry around the vertical axis, let $\mathbf{R}_{\text{sym}}$ denote this symmetry rotation, giving the orientation criterion:
\begin{equation}
\min\bigl(\| \boldsymbol{\phi}(\mathbf{R}_{\text{box}}^\top \mathbf{R}_{\text{target}}) \|_2, \|\boldsymbol{\phi}(\mathbf{R}_{\text{box}}^\top \mathbf{R}_{\text{sym}} \mathbf{R}_{\text{target}})\|_2\bigr) < \delta_\theta.
\end{equation}
\end{definition}

\subsection{Experimental Protocol}
Upon arrival, participants were introduced to the task, system, and experimental procedure via an instructional video that showed the task objective, robot control interface, and experimental flow. Participants were instructed that holding the clutch button on the VR controller would move the impedance target, while releasing the button decoupled hand motion from the robot. They were also told that forces could be applied by slightly moving the controller beyond the point of contact once the robot touched the box. If the box flipped, participants could reposition it upright with their hands, but were not allowed to move it closer to the target. After the introduction, participants signed an informed consent form and completed a demographic questionnaire. This was followed by the experimental tasks.

Each participant completed both interface conditions, with order counterbalanced across participants. Each condition included an unrecorded practice phase (4 sliding targets and 4 lifting targets) followed by the evaluation phase of 8 sliding targets and 8 lifting targets. Based on the pilot study, we chose the number of targets so that the full experiment would last roughly 30 minutes per participant to limit fatigue effects. Participants were informed about the blue disks during the first practice trial under the visualization condition. Target locations were identical across participants and conditions. Although participants seeing the same targets in both conditions might speed up performance due to memorization, we judged that using different targets between conditions would introduce larger order effects.

\subsection{Measures}
Task performance was measured using completion time, defined as the time from target presentation to satisfaction of the position and orientation criteria. After each condition, participants completed the System Usability Scale (SUS)~\cite{Brooke1996a} and the NASA Task Load Index (NASA-TLX)~\cite{Hart2006} to assess perceived usability and workload. The SUS consisted of 10 questions rated on a 5-point Likert scale, while the NASA-TLX included 6 questions rated on a scale from 0 to 10. Informal qualitative feedback from participants was paraphrased and noted by the experimenter.

\subsection{Participants}
The study involved 18 participants (6 women, 12 men), aged 18--35 (6 in the 18--24 group, 12 in the 25--35 group). Participants were divided into two groups of 9 to counterbalance visualization condition order. One participant was excluded because their session lasted more than three times the planned 30-minute duration, raising concern about fatigue effects. Data from the remaining 17 participants were included in the analysis. We conducted informal pilot tests with 10 additional participants prior to the main study to refine visualization design and experimental protocol. 

\section{RESULTS AND DISCUSSION}
This section presents the user study results and discusses observed effects of the impedance visualization. The dataset used for this analysis is  available online~\cite{VanDenBrandt2026}. We first analyze target completion times, followed by subjective measures of usability and workload, and finally qualitative feedback.

\subsection{Target Completion Time}
To examine the effect of visualization conditions on target completion time, a linear mixed-effects model (LMM) was fitted. The LMM included visualization condition, task type, and visualization order as fixed effects, along with their two-way and three-way interactions, and included participants and box targets as random intercepts. Target completion times were log-transformed to address positive skew. Estimated marginal means (EMMs) were computed for each visualization condition within each task. Pairwise contrasts of EMMs were evaluated using t-tests with Satterthwaite-adjusted degrees of freedom.

\begin{figure}[t]
    \centering
    \vspace{4pt}
    \includegraphics[]{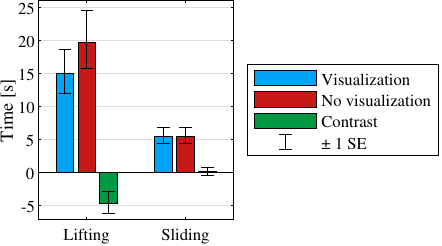}
    \vspace{-3pt}
    \caption{Target completion times for the lifting and sliding tasks, estimated from a linear mixed-effects model fitted on participant data. Reported values represent EMMs for both visualization conditions and the contrast, back-transformed from the model's log scale. }
    \label{fig:results}
\end{figure}

Fig. \ref{fig:results} shows the completion time EMMs, back-transformed from the log scale, along with the estimated contrast between visualization conditions for each task. The AR visualization significantly reduced completion times by 24\% for the lifting task ($t(507) = -2.49$, $p = 0.01$, $d = -0.30$). No significant effects were found for the sliding task ($t(507) = 0.08$, $p = 0.94$, $d = 0.00$).

These results align with the intended purpose of the visualization. Lifting tasks require precise force regulation to avoid dropping the object, making feedback on the impedance target particularly valuable. The visualization likely helped participants monitor and regulate contact forces more accurately, allowing them to establish a secure grasp more quickly and reduce drop rates. Sliding tasks, where dropping the box or establishing a grasp are not a problem, were largely unaffected, suggesting that visualization did not impair performance.

\subsection{System Usability and Perceived Workload}
To complement the objective performance metric, participants' responses to the NASA-TLX and SUS questionnaires were analyzed. SUS Question 3 was not recorded due to a technical error.

Table \ref{tab:overall} shows the aggregated NASA-TLX and SUS scores. Paired t-tests indicated no significant differences between visualization conditions for perceived workload ($t(16)=-0.39$, $p = 0.70$, $d=-0.09$) or system usability ($t(16)=1.02$, $p = 0.32$, $d=0.25$). Analysis of individual questionnaire items revealed no statistically significant effects after applying Holm--Bonferroni correction to account for multiple comparisons. 

\begin{table}[!t]
\centering
\caption{NASA-TLX and SUS scores: means and standard errors under both visualization conditions, with paired t-test results.}
\label{tab:overall}
\begin{tabular}{l|
                S[table-format=.2] S[table-format=1.2]|
                S[table-format=2.2] S[table-format=1.2]|
                S[table-format=1.2] S[table-format=0.2] S[table-format=1.2]}
\hline
         & \multicolumn{2}{c|}{Visualization} & \multicolumn{2}{c|}{No Vis.} & \multicolumn{3}{c}{t-test} \\ \hline
Survey & {Mean} & {SE} & {Mean} & {SE} & {$t(16)$} & {$p$} & {$d$} \\ \hline
TLX     & 4.34 & 0.34 & 4.40 & 0.35 & -0.39 & 0.70 & -0.09 \\
SUS     & 59.41 & 3.16 & 56.47 & 3.82 & 1.02 & 0.32 & 0.25 \\ \hline
\end{tabular}
\end{table}
These results indicate that, while the impedance visualization significantly improved objective performance in the lifting task, it did not significantly affect perceived system usability or workload. This discrepancy may reflect a higher variability of subjective measures, requiring larger samples to detect subtle effects. Another explanation could be that completion times were analyzed separately for lifting and sliding tasks, whereas the questionnaires captured overall impressions. Task-specific benefits may therefore be obscured in the aggregated ratings. 

\subsection{Qualitative feedback}
Participants provided informal feedback on the AR impedance visualization, offering insights beyond the quantitative measures. Five participants explicitly reported using the target visualization to regulate force, and two noted that the visualization was more helpful for lifting than for sliding, consistent with the patterns observed in the target completion times. Three participants additionally mentioned that the visualization assisted them in rotating the end effectors to achieve planar contact with the box. One participant initially thought the robot was overshooting the target without the visualization. Using the impedance visualization, they realized they moved the target too fast for the robot to keep up, and that the perceived overshoot was caused by moving the target too far. This suggests that the visualization could be even more beneficial for less responsive systems, such as those without velocity tracking.

Eight participants commented that the AR targets occasionally occluded their view of the robots or the box, although this did not appear to significantly impact the quantitative measures. Some participants suggested making the targets smaller or more transparent to reduce visual clutter. One participant indicated that the blue disks were helpful for understanding how the system works, but that once they became familiar with the interaction, they preferred to have the disks removed to see more clearly. This observation points to a potential order-related effect, where the visualization aids initial familiarization with impedance control, providing insights that remain valuable even after the AR targets are removed. While the current experimental design does not allow detailed analysis of such learning effects, these results suggest that the AR visualization could be valuable for training.

\section{CONCLUSION \& FUTURE WORK}
This work investigated how augmented reality can improve transparency in VR-based teleoperation with impedance control. We introduced an AR visualization that exposes the otherwise hidden impedance target pose and its positional offset, enabling users to more directly interpret the relationship between their inputs, robot motion, and resulting interaction forces. Our controlled user study ($n=17$) demonstrated that the visualization significantly improves performance in force-critical lifting tasks, reducing completion time by 24\%, while not affecting sliding tasks. Subjective measures reveal that the increase in speed is not paired with a statistically significant change in perceived workload or usability. Qualitative feedback suggests this arises because the visualization supports force regulation, helps align end-effector orientation, and prevents target overshoot.
 
These results open several directions for future research. First, investigating how the visualization performs across different robotic platforms and task domains could provide further insight into its generality. While this work focuses on one particular setup with lifting and sliding tasks, the underlying principle can extend to other manipulators and tasks. Second, future work could explore alternative designs that vary in size, translucency, or form to reduce visual occlusion. Third, a dedicated study on learning effects could examine whether the visualization builds intuition that persists after AR is removed. Finally, complementing completion time with further objective measures such as force profiles, motion trajectories, or drop rates could provide deeper insight into how the visualization influences teleoperation performance.

Overall, our findings indicate that making the impedance target visible through AR is a viable approach to improve human-robot interaction in contact-rich teleoperation without requiring additional haptic hardware.

\section*{ACKNOWLEDGEMENTS}
Generative AI was used to assist with manuscript editing.
% \clearpage
\section*{REFERENCES}
\printbibliography[heading=none]

\end{document}